\documentclass[sigconf,nonacm]{acmart}

\AtBeginDocument{%
}

\usepackage{graphicx}
\usepackage{booktabs}
\usepackage{multirow}
\usepackage{makecell}
\usepackage{pifont}
\usepackage{enumitem}
\usepackage{float}

\newcommand{\cmark}{\ding{51}}
\newcommand{\xmark}{\ding{55}}
\newcommand{\nmark}{--}

\renewcommand{\shortauthors}{Yan et al.}

\begin{document}

\title{TerraLogic: A Benchmark for Hierarchical Geospatial Reasoning in Earth Observation}

\author{Yuhang Yan}
\affiliation{%
  \institution{The Chinese University of Hong Kong (Shenzhen)}
  \country{China}
}
\email{yuhangyan@link.cuhk.edu.cn}

\author{Linchao Mou}
\affiliation{%
  \institution{Technical University of Munich}
  \country{Germany}
}
\email{lichao.mou@tum.de}

\author{Bokang Yang}
\affiliation{%
  \institution{The Chinese University of Hong Kong (Shenzhen)}
  \country{China}
}
\email{bokangyang@link.cuhk.edu.cn}

\author{Qingyu Li}
\authornote{Corresponding author.}
\affiliation{%
  \institution{The Chinese University of Hong Kong (Shenzhen)}
  \country{China}
}
\email{liqingyu@cuhk.edu.cn}

\maketitle

\noindent
{\bfseries\large Abstract}

\vspace{0.3em}

Beyond perception, reasoning is essential in remote sensing for advanced
interpretation, inference, and decision-making. Recent advances in large
language models (LLMs) have enabled tool-augmented agents that leverage
external tools to perform complex analytical tasks. However, existing
studies in remote sensing primarily focus on perception-oriented tasks,
leaving cognitive geospatial reasoning largely underexplored. To address
this gap, we introduce \textbf{TerraLogic}, a benchmark for geospatial
reasoning. TerraLogic comprises 545 scenario-driven, hierarchy-aware
tasks---such as hazard vulnerability assessment, urban heat island
analysis, and forest fragmentation dynamics---spanning optical,
Synthetic Aperture Radar (SAR), and infrared (IR) imagery. It advances
evaluation beyond recognition and monitoring toward cognitive-level
geospatial analysis. To facilitate evaluation on TerraLogic, we further
propose \textbf{HieraPlan}, a tool-augmented agent that organizes
toolkits into functional hierarchies and performs fault-tolerant
reasoning. HieraPlan enables structured abstraction, robust recovery
from tool failures, and stable long-horizon planning. Extensive
experiments demonstrate that current approaches struggle with
hierarchical geospatial reasoning, while HieraPlan provides a strong
baseline with improved reasoning, cross-modal generalization, and error
handling. The dataset and agent code are publicly available at
\url{https://github.com/Ireliya/TerraLogic}.

\vspace{0.8em}

\noindent
{\bfseries\large Keywords}

\vspace{0.3em}

\noindent
Benchmark dataset; Geospatial reasoning; Intelligent agents;
Remote sensing; Multimodal reasoning; Tool-augmented agents.

\vspace{0.8em}


\section{Introduction}

Large language models (LLMs), empowered by generative pretraining and instruction tuning, have substantially improved zero-shot task completion across diverse applications~\cite{yang2024harnessing, zhou2024larger}. Building on this progress, LLM-driven agents can decompose high-level goals into sub-tasks and orchestrate external tools for multi-step problem solving~\cite{zhao2024expel, li-2025-review}. However, extending such agents to remote sensing (RS) remains challenging. RS imagery is highly heterogeneous across modalities such as optical, Synthetic Aperture Radar (SAR), and infrared, while targets vary drastically in scale, orientation, and spatial configuration. These properties often challenge the assumptions of generic visual-language reasoning, causing agent performance to degrade under RS-specific data distributions and stricter constraints on spatial units and tool parameters.
\begin{figure}[H]
  \centering
  \includegraphics[width=\linewidth]{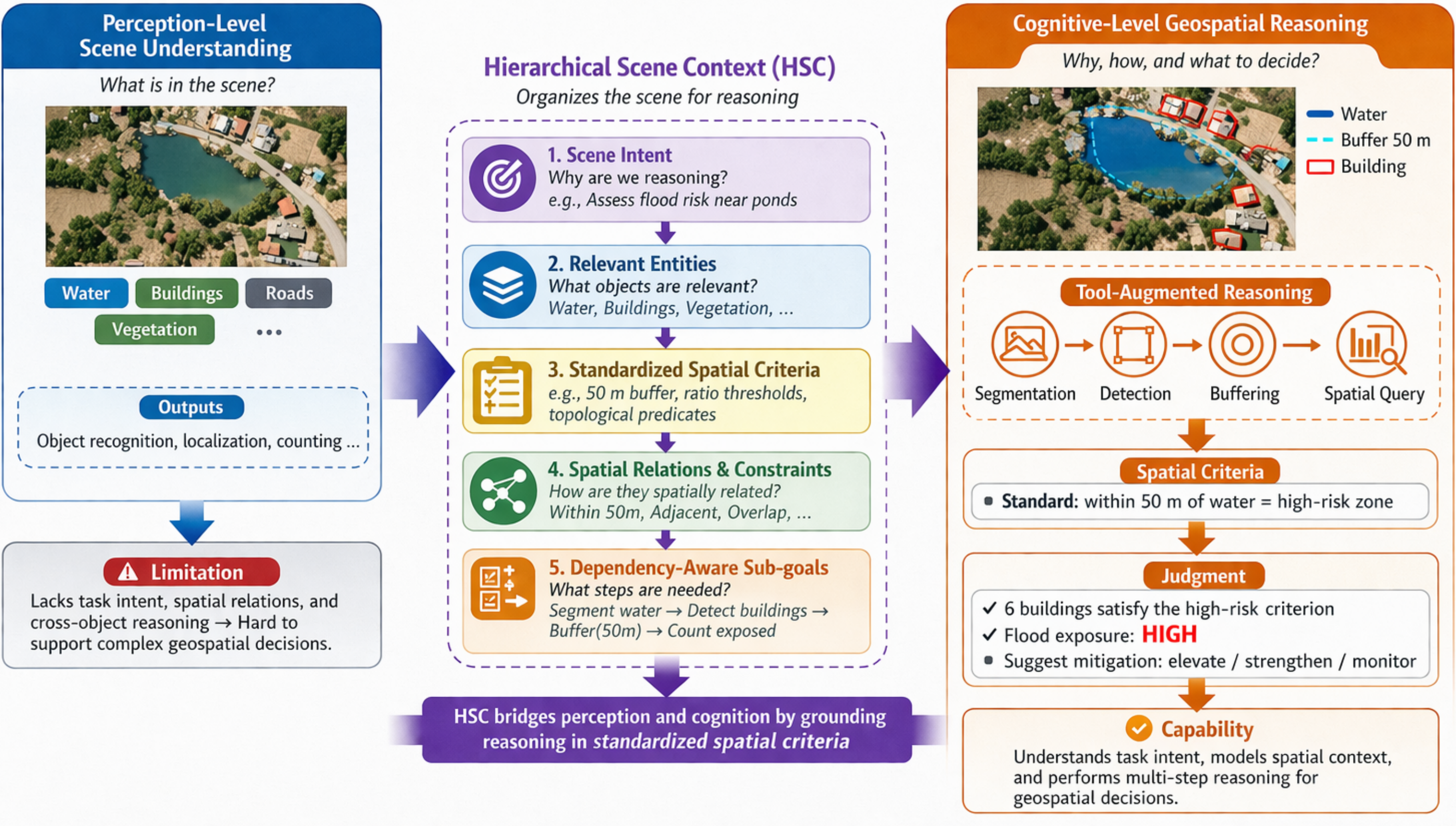}
  \caption{Hierarchical Scene Context (HSC) enables cognitive-level geospatial reasoning by organizing scene information from intent to entities, relations, and dependency-aware sub-goals, bridging perception and decision-making.}
  \label{fig:pipeline2}
\end{figure}
More importantly, real-world Earth observation (EO) applications require more than perception. Beyond perception-oriented tasks such as classification, localization, counting, and RS Visual Question Answering (VQA), many EO applications demand \emph{cognitive-level geospatial reasoning} for decision-making. These tasks are inherently scenario-driven: they require identifying task-relevant entities, composing intermediate spatial analyses, and deriving conclusions that are actionable under application context. For example, flood-risk assessment screens buildings near water bodies~\cite{oubennaceur2019flood}, irrigation analysis quantifies cropland adjacency to water sources~\cite{fu2022critical}, and ecological stabilization identifies barren patches fully enclosed by forests~\cite{hansen2009quantifying}. In such cases, perception is only the entry point; the ultimate goal is decision-oriented geospatial reasoning. What distinguishes EO reasoning from generic multi-step perception is that it must satisfy standardized geospatial criteria specified in domain guidelines, such as buffer distances, adjacency rules, ratio thresholds, or topological predicates. These criteria cannot be reliably applied from perceptual outputs alone. Their interpretation depends on \emph{why} the scene is observed, \emph{which} entities are relevant, and \emph{how} intermediate spatial products should be generated and reused. EO reasoning is therefore not merely a sequence of perception outputs and tool calls. Instead, it requires an explicit task structure that connects scenario intent, dependency-aware sub-goals, and standards-grounded spatial constraints, as illustrated in Figure~\ref{fig:pipeline2}.

Recent studies have adapted LLMs into RS agents, including RS-ChatGPT~\cite{guo2024remote}, RS-Agent~\cite{rs-agent}, GeoLLM-Engine~\cite{singh2024geollm}, Change-Agent~\cite{change-agent}, and Tree-GPT~\cite{du2023tree}. These systems show promise on perception-oriented tasks such as classification, localization, counting, and RS VQA. Recent benchmarks have begun to extend evaluation with execution-grounded tool use beyond perception outputs. ThinkGeo~\cite{thinkgeo} studies execution-grounded multi-step tool-use trajectories for RS tasks, while Earth-Agent~\cite{earthagent} emphasizes broad EO resource integration and large-scale tool orchestration. However, these advances still mainly improve task execution and tool coordination, rather than explicitly modeling three ingredients central to cognitive geospatial reasoning: \emph{(i) scenario-level intent}, \emph{(ii) dependency-aware sub-goals}, and \emph{(iii) standards-grounded spatial constraints}. As a result, tool-executed reasoning trajectories may appear plausible, yet remain non-reproducible, non-verifiable, and difficult to compare across methods.
Taken together, these limitations indicate that a benchmark for cognitive geospatial reasoning should satisfy three requirements. First, it should provide \emph{hierarchy-structured task specifications} that connect scenario intent to dependency-aware sub-goals, rather than reducing reasoning to flat tool-call sequences. Second, it should support \emph{standards-grounded and verifiable spatial reasoning}, where predicates, thresholds, and topology constraints can be checked through executable tools. Third, it should ensure \emph{cross-modality RS coverage} for robust evaluation across heterogeneous EO data, including optical, SAR, and infrared imagery. Existing benchmarks satisfy these properties only partially and often in isolation. As summarized in Table~\ref{tab:benchmark_comparison}, prior benchmarks typically support verifiable tool execution and step-level evaluation, but rarely formulate reasoning around scenario-level intent, dependency-aware sub-goals, or standards-based spatial criteria. Moreover, cross-modal reasoning remains only partially evaluated.

To address this gap, we introduce \textbf{TerraLogic}, a benchmark for cognitive-level geospatial reasoning across multiple RS modalities. TerraLogic comprises 545 scenario-driven tasks spanning optical, SAR, and infrared imagery. Each task is specified in a structured form that couples \emph{standards-anchored spatial triggers} (e.g., quantitative thresholds or topological predicates) with expert interpretation and verifiable tool chains, while organizing reasoning under scenario intents and dependency-aware sub-goals. Building on TerraLogic, we further propose \textbf{HieraPlan}, an LLM-driven agent that hierarchically organizes analytic tools and executes constraint-aware, fault-tolerant reasoning pipelines with recovery from intermediate failures.
Our main contributions are summarized as follows:
\begin{itemize}[topsep=2pt, itemsep=2pt]
    \item We present \textbf{TerraLogic}, the first benchmark for hierarchical geospatial reasoning, comprising 545 tasks across optical, SAR, and infrared modalities.
    \item We propose \textbf{HieraPlan}, an agentic framework that enables cognitive reasoning with hierarchical planning, robust error handling, and support for multi-modal RS scenarios.
    \item We establish new baselines by comprehensively evaluating state-of-the-art LLMs on \textbf{TerraLogic}, revealing both their strengths and limitations in cognitively demanding geospatial reasoning.
\end{itemize}
\begin{table}[t]
\centering
\scriptsize
\setlength{\tabcolsep}{2.5pt}
\renewcommand{\arraystretch}{0.95}

\caption{Comparison of representative \textbf{execution-grounded remote sensing (RS/EO) reasoning benchmarks}. 
(\cmark: fully supported; \xmark: not supported; \nmark: not emphasized or not applicable.)}
\label{tab:benchmark_comparison}
\vspace{2pt}

\resizebox{\columnwidth}{!}{%
\begin{tabular}{lcccccc}
\toprule
\multirow{2}{*}{Benchmark}
& \multicolumn{3}{c}{\makecell[c]{Multi-sensor}} 
& \multirow{2}{*}{\makecell[c]{Verifiable\\execution}}
& \multirow{2}{*}{\makecell[c]{Step-level\\eval.}}
& \multirow{2}{*}{\makecell[c]{Hierarchical\\scene context}} \\
\cmidrule(lr){2-4}
& Optical & SAR & IR & & & \\
\midrule
UnivEARTH~\citep{univearth}
& \cmark
& \xmark
& \cmark
& \cmark
& \nmark
& \nmark \\
GeoLLM-Engine~\citep{singh2024geollm}
& \cmark
& \cmark
& \xmark
& \cmark
& \cmark
& \nmark \\
ThinkGeo~\cite{thinkgeo}
& \cmark
& \cmark
& \xmark
& \cmark
& \cmark
& \nmark \\
Earth-Bench~\cite{earthagent}
& \cmark
& \xmark
& \xmark
& \cmark
& \cmark
& \nmark \\
\textbf{TerraLogic (Ours)}
& \cmark
& \cmark
& \cmark
& \cmark
& \cmark
& \cmark \\
\bottomrule
\end{tabular}%
}
\end{table}
\section{TerraLogic Benchmark Construction}
We propose \textbf{TerraLogic}, a benchmark designed to assess the geospatial reasoning capabilities of tool-augmented agents powered by LLMs. TerraLogic integrates diverse imagery from optical, SAR, and infrared modalities with expert-curated knowledge and tool-augmented query pipelines, yielding 545 high-quality instances.

\begin{figure*}[!t]
  \centering
  \includegraphics[width=0.75\textwidth]{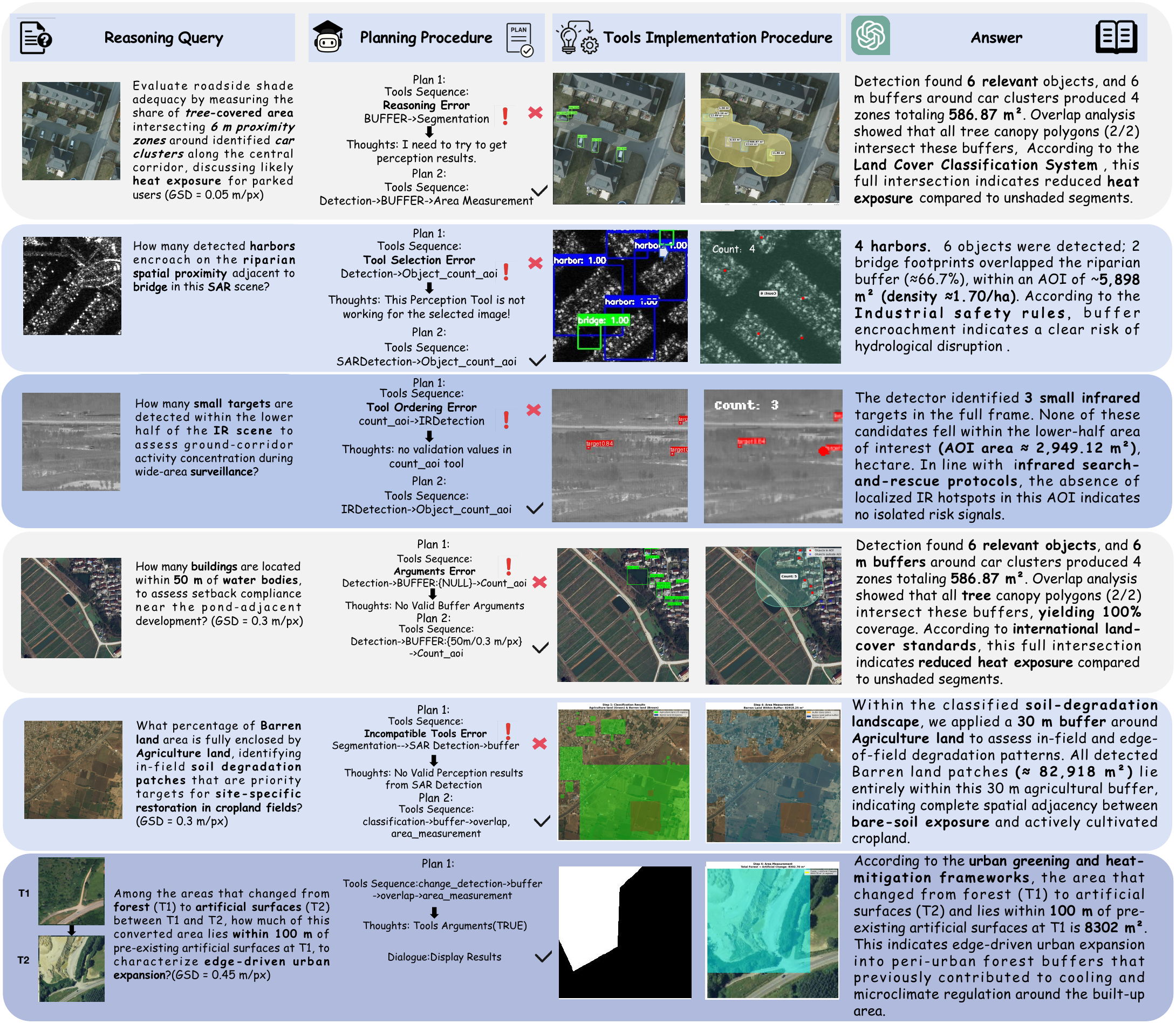}
  \caption{Representative TerraLogic tasks and HieraPlan execution traces under our error taxonomy.
  From left to right: (\emph{i}) input image and query, (\emph{ii}) high-level plans from HieraPlan,
  (\emph{iii}) executed tool sequence, and (\emph{iv}) final geospatial answer. Red crosses mark erroneous
  intermediate plans or tool calls, while green checkmarks indicate the validated final tool chain and answer.}
  \label{fig:datasets}
\end{figure*}

Each instance couples real-world imagery with structured, multi-step reasoning challenges that require both low-level perception (e.g., segmentation, detection) and high-level decision-making (e.g., urban planning, disaster assessment). Unlike prior remote sensing benchmarks that focus narrowly on perception (e.g., classification or detection), TerraLogic emphasizes \emph{end-to-end reasoning}, from perception through spatial analysis to actionable conclusions, thereby providing a rigorous testbed for agents. Figure \ref{fig:datasets} illustrates six representative samples from the TerraLogic benchmark.

\subsection{Geospatial Reasoning Scenarios}

A central challenge in RS is bridging the gap between low-level perceptual outputs and the high-level intelligence required for real-world decision-making. To address this, we adopt a structured taxonomy of geospatial reasoning tasks, systematically organized into seven primary domains: urban planning, disaster assessment, environmental monitoring, transportation analysis, activation monitoring, maritime monitoring, and industrial sites.

\subsection{Source Data Construction Pipeline}

\begin{figure*}[t]
  \centering
  \includegraphics[width=0.8\textwidth]{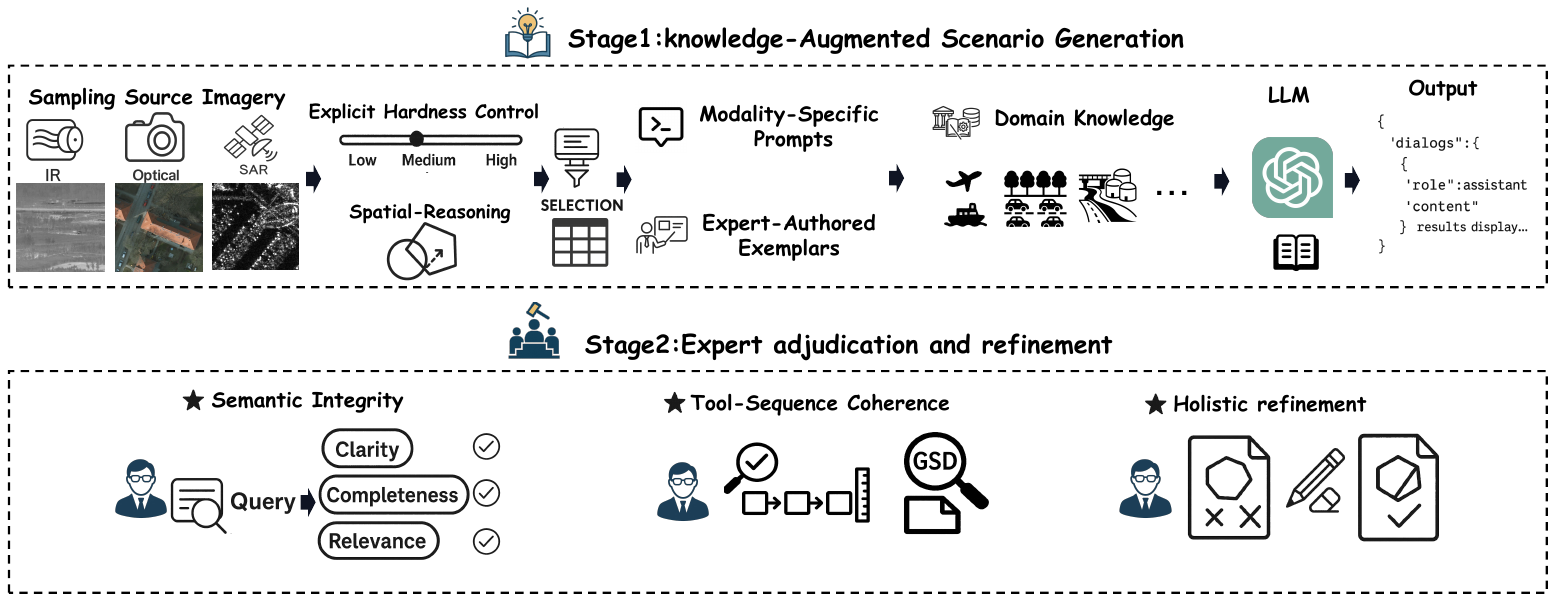}
  \caption{Pipeline for constructing the TerraLogic benchmark.}
  \label{fig:pipeline}
\end{figure*}

We construct TerraLogic with a two-stage pipeline (see Figure \ref{fig:pipeline}) that combines diverse source imagery, knowledge-augmented scenario generation, and multi-pass expert adjudication to produce 545 validated instances. The source datasets used in this pipeline are summarized in Table~\ref{tab:datasets}.

\subsubsection{Stage 1: Knowledge-augmented scenario generation.}

We obtain a stratified sample of source imagery (by modality and scene type) and apply explicit hardness controls to select cases that require multi-tool reasoning and compositional spatial relations (e.g., proximity, containment, topology). Candidate queries and tool-chains are generated with ChatGPT-5 via in-context learning: prompts are modality-specific and seeded with expert-authored exemplars.

To ground generation in domain knowledge, we inject a compact knowledge corpus into prompts. The corpus covers four guidance categories: (i) urban greening and heat-mitigation frameworks \citep{twohig2018health,20,30,50,r50}; (ii) international land-cover standards \citep{51,52}; (iii) aviation and maritime search-and-rescue doctrine for IR small-target tasks \citep{s1}; and (iv) industrial safety rules for separation and proximity of hazardous assets \citep{safety,safety2}. These anchors force the ChatGPT-5 to produce quantitative thresholds (distances, areas, class compositions) consistent with established frameworks. We also enforce operator legality checks (including ground distance sampling (GSD)-aware units) to prevent invalid tool sequences.

\subsubsection{Stage 2: Expert adjudication and quality gating.}

We perform expert-led curation with an \emph{execution-gated} release criterion. 
Specifically, each Stage~1 candidate is subjected to a three-round vetting procedure by eight remote-sensing experts, aiming to produce instances that are both unambiguous in intent and verifiable through tool execution. \textit{Checklist-based assessment.}
For every candidate, reviewers examine three aspects: 
(i) \emph{semantic soundness} (the query states a well-posed, domain-relevant task without underspecification), 
(ii) \emph{workflow validity} (the operator choices, ordering, and arguments comply with modality constraints and remain consistent under GSD-aware units), and 
(iii) \emph{geospatial correctness} (no geometry/topology pitfalls, and intermediate spatial objects are meaningful for downstream operators). 
We iteratively revise each candidate until all criteria are satisfied; otherwise, we discard it if the intended question cannot be substantiated by executable tool evidence (invalid units, ill-defined spatial predicates, or inconsistencies between the query and the expected tool outputs).

\textit{Execution-gated verification.}
To avoid releasing instances that rely on free-form generation, we require that the finalized tool sequence can be \emph{run to completion} under operator legality constraints (typed I/O, unit validity, and geospatial constraints). 
We accept an instance only when (a) the tool chain executes without violations, (b) intermediate outputs pass basic validity checks (non-empty outputs where required and valid topology when applicable), and (c) any reported quantities in the final answer are directly computed from tool outputs rather than inferred without evidence. 
This gating criterion ensures that every released instance is paired with an executable tool chain that supports traceable, evidence-grounded evaluation.

\textit{Rounds and traceability.}
The three rounds consist of an initial screening, disagreement resolution for borderline cases, and a final conformance check across the query text, tool arguments, and produced outputs. 
All modifications to the query, operator selection, arguments, and final answer are logged in the curation interface, preserving instance-level traceability for verification and subsequent analysis.

\begin{table}[t]
\centering
\small
\setlength{\tabcolsep}{4pt}
\renewcommand{\arraystretch}{1.1}

\caption{Datasets used as image sources in the TerraLogic benchmark.}
\label{tab:datasets}

\resizebox{\columnwidth}{!}{%
\begin{tabular}{l l c c l}
\hline
\textbf{Name} & \textbf{Annotation Type} & \textbf{Resolution} & \textbf{Modality} & \textbf{Geographical Coverage} \\
\hline
LoveDA~\citep{wang2021loveda}           & Masks          & 0.3 m/px  & Optical & Nanjing/Changzhou/Wuhan, China \\
ISPRS Potsdam~\citep{potsdam}           & Masks          & 0.05 m/px & Optical & Potsdam, Germany \\
ISPRS Vaihingen~\citep{isprs_vaihingen} & Masks          & 0.09 m/px & Optical & Vaihingen, Germany \\
DeepGlobe Land Cover~\citep{deepglobe}  & Masks          & 0.50 m/px & Optical & Cities in India/Indonesia/Thailand \\
HRSCD~\citep{hrscd}                     & Change masks   & 0.45 m/px & Optical & Rennes/Caen, France \\
OGSOD~\citep{ogsod}                     & Bounding boxes & 3 m/px    & SAR     & Cities in China \\
DMIST~\citep{dmist}                     & Bounding boxes & --        & IR      & Chengdu, China \\
\hline
\end{tabular}%
}
\end{table}

\subsection{Benchmark Statistics}

TerraLogic contains 545 queries across optical, SAR, and IR modalities. In total, 13 tools are invoked 1649
 times, with most queries requiring multi-tool composition. Query lengths range from short factual questions to complex compositional ones, demonstrating both diversity and reasoning depth. Detailed statistics are reported in Table~\ref{tab:question_stats}. Each reasoning task in TerraLogic is instantiated as a distinct scene context type, characterized by three components: (i) a \textsc{Spatial Trigger} (quantitative thresholds or topological relations), (ii) an \textsc{Expert Interpretation} (domain-grounded semantics), and (iii) a \textsc{Verifiable Tool-Chain}. This hierarchical design bridges perception and cognition, offering structured priors and explicit multi-step pathways while ensuring interpretability and verifiability. By covering both canonical and high-complexity reasoning scenarios, TerraLogic provides rigorous testbeds for evaluating diverse agentic capabilities, including fine-grained spatial understanding, multi-step tool composition, quantitative analysis, and context-aware risk assessment.

\begin{table}[t]
\centering
\caption{Statistics of question complexity and tool-use patterns across different modalities.}
\label{tab:question_stats}
\footnotesize
\renewcommand{\arraystretch}{1.1}
\setlength{\tabcolsep}{4pt}

\begin{tabular}{lcccc}
\hline
\textbf{Metric} & \textbf{All} & \textbf{Optical} & \textbf{SAR} & \textbf{IR} \\
\hline
Total queries & 545 & 386 & 116 & 43 \\
Total tool calls & 1649 & 1217 & 346 & 86 \\
2/3/4-tools use & 72/387/86 & 27/273/86 & 2/114/0 & 43/0/0 \\
Maximum question length & N/A & 406 & 207 & 288 \\
Minimum question length & N/A & 109 & 111 & 135 \\
Average question length & 158.24 & 160.13 & 143.49 & 181.05 \\
\hline
\end{tabular}
\end{table}

\section{The HieraPlan Agent}

To support evaluation on \textbf{TerraLogic}, we provide \textbf{HieraPlan}, a hierarchy-aware and fault-tolerant agent for geospatial reasoning in remote sensing. Given a task, HieraPlan retrieves a task-relevant subset of tools from a hierarchical library spanning \emph{Perception}, \emph{Spatial-Relations}, \emph{Spatial-Statistics}, \emph{SAR}, and \emph{IR}, then generates a modality-aware plan, executes concrete operators, and synthesizes the final answer from validated tool outputs. HieraPlan supports \emph{hierarchical abstraction}: tools are organized by function and modality, with typed inputs/outputs and basic legality constraints, reducing invalid compositions over heterogeneous RS data. Its executable operators are grounded in remote-sensing expert models, including \textbf{RemoteSAM}~\citep{yao2025remotesam}, \textbf{SARATR-X}~\citep{sar1,sar2}, \textbf{DMIST}/\textbf{LASNet}~\citep{dmist}, and \textbf{Change3D}~\citep{change3d}. It further enables \emph{fault-tolerant execution}: when a step fails, the agent first attempts within-toolkit substitution and, if needed, replans only the remaining suffix while preserving the validated prefix. Finally, it enforces \emph{verifiable grounding}: quantitative outputs are computed by tools, while the LLM is restricted to natural-language synthesis from validated intermediate results. Overall, HieraPlan serves as a suitable reference baseline for TerraLogic, whose tasks require scenario-aware decomposition, cross-modal tool coordination, and reliable execution under long reasoning chains.

\section{Experiments and Results} 
To evaluate the reasoning and tool-use capabilities of \textbf{HieraPlan} under real-world RS scenarios, we conduct comprehensive experiments on the \textbf{TerraLogic} benchmark. Since HieraPlan is controlled by an LLM, we evaluate a diverse set of foundation models, including API-based systems---GPT-3.5~\citep{gpt-3.5}, GPT-4o~\citep{4o}, and Gemini-2.5-Flash~\citep{team2023gemini}---as well as a broad range of open-source LLMs. The open-source models include large-capacity systems, such as DeepSeek-V3~\citep{bi2024deepseek}, Qwen2.5-32B-Instruct~\citep{team2024qwen2}, and LLaMA-3-70B-Instruct~\citep{LLaMA-3-8B}, together with medium-sized models, including Qwen2.5-7B~\citep{qwen7b}, InternLM3-8B~\citep{InternLM3-8B}, LLaMA-3-8B~\citep{LLaMA-3-8B}, Mistral-7B~\citep{Mistral-7B(-Instruct)}, Yi-1.5-6B~\citep{Yi-1.5-6B}, and Phi-3-Mini~\citep{abdin2024phi}. All experiments are conducted on an NVIDIA A5000 GPU within the OpenCompass~\citep{opencompass} evaluation platform.
\subsection{Overall Performance}

We evaluate \textbf{HieraPlan} on the \textbf{TerraLogic benchmark} under a ReAct-style~\citep{yao2023react} protocol with both \emph{step-by-step} and \emph{end-to-end} settings. The step-by-step metrics---InstAcc, ToolAcc, ArgAcc, and SummAcc---measure instruction following, tool selection, argument correctness, and summary generation, respectively. The end-to-end scores---P, O, and L---assess tool-augmented predictions on the \emph{Perception}, \emph{Spatial-Statistics}, and \emph{Spatial-Relations} toolkits. Within this setup, HieraPlan complements rather than alters the ReAct protocol, providing hierarchical orchestration while preserving the same evaluation criteria. For \emph{final answer accuracy} (AnsAcc), we avoid deterministic string matching (e.g., GTA~\citep{wang2024gta}), which is brittle to minor lexical variations and may penalize semantically correct outputs. This is especially important in TerraLogic, where final answers are synthesized from \textit{query semantics, tool inputs/outputs, and scenario knowledge}, rather than matched to a single canonical string. Following prior work on \textbf{LLM-as-a-judge}~\cite{judge1,judge2,judge3}, we compute \textbf{AnsAcc} using GPT-4o-mini with a rubric-based prompt to assess whether the predicted answer and its key arguments are semantically consistent with the ground truth. Table~\ref{tab:geoplanner_results} reports the overall results on TerraLogic, whose multimodal, scenario-driven tasks require agents to coordinate multiple \textbf{toolkits} spanning perception, spatial analysis, SAR, and IR, thereby enabling systematic evaluation of cognitive-level geospatial reasoning. As shown in Table~\ref{tab:geoplanner_results}, \textbf{GPT-4o} achieves the strongest overall performance across both evaluation settings, leading on InstAcc, ArgAcc, SummAcc, as well as P, O, and Ans. Among open-source large models, \textbf{Llama-3-70B-Instruct} performs best overall, with competitive ToolAcc and strong end-to-end results. Among medium-scale open-source models, \textbf{Qwen2.5-7B-Instruct} is notable for achieving the highest L score (\emph{Spatial Relations}) across all evaluated models, despite weaker InstAcc and SummAcc. Across model families, \textbf{ArgAcc} remains the main bottleneck, while lower ToolAcc is consistently associated with reduced end-to-end performance, highlighting the difficulty of reliable tool selection and faithful argument propagation in long-horizon geospatial workflows.

\begin{table}[t]
\centering
\small
\caption{\textbf{Main results of HieraPlan on the TerraLogic benchmark.}
Inst., Tool., Arg., and Summ.\ denote step-by-step accuracies (InstAcc, ToolAcc, ArgAcc, SummAcc).
P, O, and L denote end-to-end scores in the \textit{Perception}, \textit{Spatial Statistics}, and \textit{Spatial Relations} toolkits.
\textbf{Ans.} denotes the final answer accuracy (AnsAcc).
\textbf{Bold} indicates the best overall score, and \underline{underline} marks the best within the same model scale.}
\label{tab:geoplanner_results}

\resizebox{\columnwidth}{!}{%
\begin{tabular}{l|cccc|cccc}
\hline
\textbf{Model} 
& \multicolumn{4}{c|}{\textbf{Step-by-Step Metrics (\%)}} 
& \multicolumn{4}{c}{\textbf{End-to-End Metrics (\%)}} \\
\cline{2-9}
& \textbf{Inst.} & \textbf{Tool.} & \textbf{Arg.} & \textbf{Summ.} 
& \textbf{P} & \textbf{O} & \textbf{L} & \textbf{Ans.} \\
\hline

\multicolumn{9}{l}{\textbf{API-based}} \\
\hline
GPT-4o
& \textbf{\underline{79.40}} & 72.01 & \textbf{\underline{23.57}} & \textbf{\underline{82.83}}
& \textbf{\underline{72.73}} & \textbf{\underline{88.38}} & 77.95 & \textbf{\underline{25.70}} \\
Gemini-2.5-Flash
& 59.09 & 43.31 & 15.75 & 47.22
& 41.47 & 40.89 & 41.51 & 12.04 \\
GPT-3.5
& 59.05 & 57.12 & 21.88 & 59.08
& 61.22 & 61.04 & 53.30 & 24.40 \\
\hline

\multicolumn{9}{l}{\textbf{Open-source (Large)}} \\
\hline
DeepSeek-V3
& 65.84 & 66.33 & 23.32 & 60.92
& 65.44 & 66.04 & 57.93 & 21.28 \\
Qwen2.5-32B-Instruct
& 62.83 & 58.00 & 17.41 & 58.90
& 44.28 & 67.40 & 61.47 & 19.08 \\
Llama-3-70B-Instruct
& 67.89 & 67.47 & 22.66 & 69.91
& 62.72 & 71.62 & 62.54 & 22.75 \\
\hline

\multicolumn{9}{l}{\textbf{Open-source (Medium/Small)}} \\
\hline
Qwen2.5-7B-Instruct
& 77.19 & \textbf{\underline{74.71}} & 23.11 & 65.32
& 68.56 & 73.03 & \textbf{\underline{82.42}} & 22.57 \\
InternLM3-8B-Instruct
& 61.26 & 57.72 & 22.59 & 50.28
& 61.44 & 53.70 & 60.70 & 18.90 \\
LLaMA3-1-8B
& 69.46 & 65.99 & 22.27 & 55.60
& 68.42 & 60.34 & 70.82 & 16.70 \\
Phi-3-Mini-4K-Instruct
& 55.45 & 41.14 & 17.17 & 10.09
& 43.63 & 7.52 & 51.74 & 3.67 \\
Mistral-7B-Instruct-v0.2
& 60.03 & 47.05 & 16.73 & 46.06
& 49.75 & 44.05 & 45.23 & 13.21 \\
Yi-1.5-6B-Chat
& 63.38 & 42.08 & 16.00 & 37.98
& 47.44 & 38.73 & 36.02 & 10.83 \\
\hline
\end{tabular}%
}
\end{table}

\subsection{Task Error Analysis.}
To explain the performance gaps above, we analyze failures using an error taxonomy covering \emph{format}, \emph{reasoning}, \emph{perception}, and \emph{tool use}. As shown in Table~\ref{tab:error_types}, \emph{format errors} are the dominant failure mode across model families. They are particularly severe for open-source models, including \textbf{DeepSeek-V3} (66.40\%), \textbf{Llama-3-70B} (55.20\%), and \textbf{Phi-3-Mini} (45.60\%), but remain substantial even for \textbf{GPT-4o} (41.65\%). This suggests that generating valid structured tool calls, especially well-formed JSON arguments, remains a major bottleneck. \emph{Tool-use errors} further distinguish model classes. \textbf{GPT-4o} is the most reliable (3.60\%), whereas \textbf{Gemini-2.5-Flash} (37.26\%) and \textbf{GPT-3.5} (39.50\%) more often select inappropriate operators or produce invalid parameters. Some open-source models show lower tool-use error rates, but this likely reflects more conservative behavior rather than stronger execution. Another major source of failure is \emph{perception grounding}. For example, \textbf{GPT-4o} shows a perception error rate of 41.26\%, indicating that interpreting land-cover masks, change maps, and spatial layouts remains challenging. By contrast, \emph{reasoning errors} are relatively moderate across models (13--35\%) and do not appear to be the main bottleneck. Overall, these results suggest that current geospatial agents are constrained more by structured output generation and multimodal grounding than by abstract reasoning alone.

\begin{table}[!t]
\centering
\small
\caption{Percentage distribution of four error categories for different LLMs (\%). \textbf{Bold} indicates the most frequent error type for each model.}
\label{tab:error_types}

\begin{tabular}{lcccc}
\hline
\textbf{Model} & \textbf{Format} & \textbf{Reasoning} & \textbf{Perception} & \textbf{Tool-use} \\
\hline
GPT-4o           & \textbf{41.65} & 13.49 & 41.26 &  3.60 \\
Gemini-2.5-Flash & 29.47 & 21.18 & 12.09 & \textbf{37.26} \\
GPT-3.5          & 31.20 & 18.60 & 10.70 & \textbf{39.50} \\
DeepSeek-V3      & \textbf{66.40} & 23.00 &  8.70 &  1.90 \\
Qwen2.5-32B      & 31.27 & 19.78 & 13.99 & \textbf{34.96} \\
Llama-3-70B      & \textbf{55.20} & 25.70 & 18.20 &  0.90 \\
Qwen2.5-7B       & \textbf{42.65} & 23.48 & 29.37 &  4.50 \\
InternLM3-8B     & \textbf{32.70} & 23.90 & 11.40 & 32.00 \\
LLaMA3.1-8B      & \textbf{38.36} & 24.58 & 20.28 & 16.78 \\
Phi-3-Mini-4K    & \textbf{45.60} & 34.70 &  1.80 & 17.90 \\
Mistral-7B-v0.2  & \textbf{45.40} & 21.20 &  5.40 & 28.00 \\
Yi-1.5-6B        & \textbf{37.90} & 28.60 &  8.60 & 24.90 \\
\hline
\end{tabular}
\end{table}

\subsection{Comparison with Remote Sensing Multimodal LLMs}

To contextualize TerraLogic’s difficulty, we compare \textbf{HieraPlan} with existing \textbf{remote-sensing multimodal LLMs (RS-MLLMs)} on the optical subset, where these models are expected to perform best. Table~\ref{tab:optical_accuracy_only} reports the results. Existing RS-MLLMs perform poorly on TerraLogic’s optical geospatial reasoning tasks: \textbf{EarthDial}, \textbf{GeoChat}, and \textbf{GeoPix} achieve only 9.38\%, 6.70\%, and 6.17\% accuracy, respectively, despite being designed for remote-sensing understanding. In contrast, \textbf{HieraPlan (GPT-4o)} reaches \textbf{29.29\%} accuracy, substantially outperforming these specialized RS-MLLMs. Nevertheless, the overall performance remains low, indicating that TerraLogic’s optical tasks are still highly challenging and require fine-grained spatial reasoning beyond standard VQA-style interpretation.

\begin{table}[!t]
\centering
\caption{Optical accuracy comparison between representative remote-sensing multimodal LLMs (RS-MLLMs) and our \textbf{HieraPlan (GPT-4o)} on TerraLogic optical VQA tasks.}
\label{tab:optical_accuracy_only}
\small
\setlength{\tabcolsep}{6pt}
\renewcommand{\arraystretch}{1.2}

\begin{tabular}{l c}
\hline
\textbf{Model} & \textbf{Optical Accuracy (\%)} \\
\hline

\multicolumn{2}{l}{\textbf{RS-MLLMs}} \\
\hline
EarthDial\citep{earthdial1}      & 9.38 \\
GeoChat\citep{geochat1}          & 6.70 \\
GeoPix\citep{geopix1}            & 6.17 \\
\hline

\multicolumn{2}{l}{\textbf{LLM-based Agent (Ours)}} \\
\hline
\textbf{HieraPlan (GPT-4o)}   & \textbf{\underline{29.29}} \\
\hline

\end{tabular}
\end{table}

\subsection{Fine-grained capability subsets.}
To pinpoint where failures arise, Table~\ref{tab:fine_grained_results} stratifies TerraLogic into four subsets with progressively increasing capability composition and planning horizon. \textbf{Semantic + Geometry (S+G)} includes tasks requiring semantic perception followed by geometric reasoning (e.g., segmentation with buffering or overlap), without quantitative statistics. \textbf{Semantic + Quantitative (S+Q)} combines semantic perception with numerical measurement (e.g., detection with distance estimation or counting), without geometric predicates. \textbf{Semantic + Geometry + Quantitative (S+G+Q)} represents full-pipeline workflows integrating all three capabilities. \textbf{Long-Horizon (H)} includes instances with $K\ge4$ sequential tool operations, and we further report \textbf{$H\cap$(S+G+Q)} to isolate the most challenging long-horizon full-pipeline cases. Two trends emerge. First, \emph{capability composition} is substantially harder than any individual capability: for most models, performance drops from S+G or S+Q to S+G+Q, indicating that errors accumulate when arguments must be propagated across heterogeneous toolkits. Second, \emph{planning horizon} remains the dominant challenge: on $H\cap$(S+G+Q), even strong systems retain moderate step-level accuracies but still degrade sharply in final answer accuracy, suggesting that small local tool or argument errors can derail end-to-end correctness in long reasoning chains.
\begin{table}[!htbp]
\centering
\scriptsize
\setlength{\tabcolsep}{2.2pt}
\renewcommand{\arraystretch}{0.86}

\caption{Fine-grained results across models and capability subsets. All metrics are in \%.}
\label{tab:fine_grained_results}

\resizebox{\columnwidth}{!}{
\begin{tabular}{l l c c c c c}
\toprule
Model & Task & Inst & Tool & Arg & Summ & Ans \\
\midrule

\multicolumn{7}{l}{\textbf{API-based}} \\
\midrule
\multirow{4}{*}{GPT-4o}
& S+G              & 77.78 & \underline{92.11} & \textbf{32.89} & 75.00 & \textbf{45.29} \\
& S+Q              & 78.33 & 89.20 & 22.80 & 94.00 & 36.00 \\
& S+G+Q            & \textbf{85.20} & \textbf{84.40} & \textbf{27.37} & \textbf{91.08} & \textbf{35.43} \\
& H $\cap$ (S+G+Q) & 60.70 & 46.77 & \underline{18.09} & 60.47 & 9.81 \\
\midrule

\multirow{4}{*}{Gemini-2.5-Flash}
& S+G              & 49.44 & 48.03 & 18.42 & 53.57 & 21.43 \\
& S+Q              & 30.67 & 25.60 & 12.00 & 30.00 & 8.00  \\
& S+G+Q            & 34.51 & 23.40 & 9.00  & 24.93 & 9.97  \\
& H $\cap$ (S+G+Q) & 50.81 & 34.24 & 12.40 & 41.86 & 0.00 \\
\midrule

\multirow{4}{*}{GPT-3.5}
& S+G              & 67.20 & 73.00 & 27.00 & 64.30 & \underline{42.90} \\
& S+Q              & \underline{83.30} & 94.40 & \underline{40.80} & \underline{98.00} & \textbf{48.00} \\
& S+G+Q            & 53.90 & 53.50 & 20.80 & 51.20 & \underline{24.70} \\
& H $\cap$ (S+G+Q) & 67.10 & \underline{54.40} & \textbf{18.30} & 69.80 & 10.50 \\
\midrule

\multicolumn{7}{l}{\textbf{Open-source (Large)}} \\
\midrule
\multirow{4}{*}{DeepSeek-V3}
& S+G              & \underline{80.00} & 86.80 & 28.30 & \textbf{92.90} & 32.10 \\
& S+Q              & \underline{83.30} & \textbf{96.80} & 30.00 & \underline{98.00} & \underline{40.00} \\
& S+G+Q            & 61.90 & 65.70 & \underline{24.00} & 53.80 & 21.30 \\
& H $\cap$ (S+G+Q) & 70.70 & \textbf{54.80} & 17.70 & 60.50 & 5.80 \\
\midrule

\multirow{4}{*}{Qwen2.5-32B-Instruct}
& S+G              & 59.40 & 65.10 & 23.70 & 60.70 & 10.70 \\
& S+Q              & \underline{83.30} & 94.40 & \textbf{41.20} & \underline{98.00} & 32.00 \\
& S+G+Q            & 60.10 & 57.50 & 15.40 & 52.00 & 21.30 \\
& H $\cap$ (S+G+Q) & 66.00 & 46.50 & 15.20 & 66.30 & 5.80 \\
\midrule

\multirow{4}{*}{Llama-3-70B-Instruct}
& S+G              & \underline{80.00} & 80.26 & 24.34 & \underline{89.29} & 25.00 \\
& S+Q              & \textbf{83.33} & 92.80 & 38.80 & 94.00 & 34.00 \\
& S+G+Q            & 64.44 & 70.12 & 23.17 & \underline{62.99} & 23.10 \\
& H $\cap$ (S+G+Q) & \underline{72.21} & 47.67 & 15.37 & \textbf{80.23} & \underline{11.63} \\
\midrule

\multicolumn{7}{l}{\textbf{Open-source (Medium/Small)}} \\
\midrule
\multirow{4}{*}{Qwen2.5-7B-Instruct}
& S+G              & \underline{80.00} & \textbf{94.70} & \underline{32.20} & 85.70 & \underline{42.90} \\
& S+Q              & \underline{83.30} & \underline{96.00} & 32.40 & \textbf{100.00} & 34.00 \\
& S+G+Q            & \underline{78.70} & \underline{77.70} & 23.50 & 57.50 & 22.00 \\
& H $\cap$ (S+G+Q) & 69.00 & 53.60 & 17.10 & 73.30 & \textbf{12.80} \\
\midrule

\multirow{4}{*}{InternLM3-8B-Instruct}
& S+G              & 74.40 & 82.90 & 28.90 & 78.60 & 25.00 \\
& S+Q              & \underline{83.30} & 94.40 & \textbf{41.20} & \textbf{100.00} & 32.00 \\
& S+G+Q            & 55.80 & 55.20 & 22.20 & 38.60 & 19.70 \\
& H $\cap$ (S+G+Q) & 70.20 & 49.60 & 16.80 & 64.00 & 5.80 \\
\midrule

\multirow{4}{*}{LLaMA3-1-8B}
& S+G              & \underline{80.00} & 89.50 & 28.30 & 85.70 & 21.40 \\
& S+Q              & 51.70 & 51.60 & 28.40 & 40.00 & 6.00 \\
& S+G+Q            & 71.20 & 71.00 & 23.20 & 50.70 & 19.20 \\
& H $\cap$ (S+G+Q) & 67.30 & 48.70 & 16.00 & \underline{76.70} & 10.50 \\
\midrule

\multirow{4}{*}{Phi-3-Mini-4K-Instruct}
& S+G              & 51.11 & 42.11 & 14.47 & 32.14 & 7.14 \\
& S+Q              & 77.67 & 82.00 & 32.40 & 72.00 & 16.00 \\
& S+G+Q            & 57.35 & 40.98 & 18.34 & 2.62  & 2.62 \\
& H $\cap$ (S+G+Q) & 41.86 & 28.29 & 8.79  & 0.00  & 0.00 \\
\midrule

\multirow{4}{*}{Mistral-7B-Instruct-v0.2}
& S+G              & 73.90 & 74.30 & 24.30 & 78.60 & 25.00 \\
& S+Q              & 77.70 & 86.00 & 36.40 & 92.00 & 16.00 \\
& S+G+Q            & 56.00 & 43.60 & 15.10 & 37.30 & 13.60 \\
& H $\cap$ (S+G+Q) & 65.30 & 40.80 & 14.30 & 47.70 & 5.80 \\
\midrule

\multirow{4}{*}{Yi-1.5-6B-Chat}
& S+G              & \textbf{81.67} & 63.82 & 23.68 & 85.71 & 14.29 \\
& S+Q              & 54.33 & 56.40 & 21.20 & 38.00 & 14.00 \\
& S+G+Q            & 60.37 & 41.02 & 15.49 & 30.97 & 11.02 \\
& H $\cap$ (S+G+Q) & \textbf{73.37} & 36.82 & 14.60 & 53.49 & 6.98 \\
\bottomrule
\end{tabular}
}
\end{table}

\section{Conclusion}

In this work, we introduced TerraLogic, the first benchmark for cognitive-level geospatial reasoning across optical, SAR, and IR modalities. TerraLogic comprises 545 hierarchy-aware, scenario-driven tasks that extend evaluation beyond perception to structured, multi-step analysis. To support benchmark evaluation, we further provide HieraPlan, a hierarchy-aware, fault-tolerant baseline agent for multimodal tool-augmented reasoning. Experiments on TerraLogic show that even frontier models remain limited, with clear weaknesses in argument prediction, summary generation, and multimodal tool integration, highlighting substantial headroom for future research on geospatial reasoning, planning, and execution.
\clearpage
\bibliographystyle{ACM-Reference-Format}
\bibliography{software}

@inproceedings{yao2023react,
  title={React: Synergizing reasoning and acting in language models},
  author={Yao, Shunyu and Zhao, Jeffrey and Yu, Dian and Du, Nan and Shafran, Izhak and Narasimhan, Karthik and Cao, Yuan},
  booktitle={International Conference on Learning Representations (ICLR)},
  year={2023}
}

@article{wang2024gta,
  title={GTA: a benchmark for general tool agents},
  author={Wang, Jize and Zerun, Ma and Li, Yining and Zhang, Songyang and Chen, Cailian and Chen, Kai and Le, Xinyi},
  journal={Advances in Neural Information Processing Systems},
  volume={37},
  pages={75749--75790},
  year={2024}
}

@article{thinkgeo,
  title={ThinkGeo: Evaluating Tool-Augmented Agents for Remote Sensing Tasks},
  author={Shabbir, Akashah and Munir, Muhammad Akhtar and Dudhane, Akshay and Sheikh, Muhammad Umer and Khan, Muhammad Haris and Fraccaro, Paolo and Moreno, Juan Bernabe and Khan, Fahad Shahbaz and Khan, Salman},
  journal={arXiv preprint arXiv:2505.23752},
  year={2025}
}

@article{change-agent,
  title={Change-agent: Towards interactive comprehensive remote sensing change interpretation and analysis},
  author={Liu, Chenyang and Chen, Keyan and Zhang, Haotian and Qi, Zipeng and Zou, Zhengxia and Shi, Zhenwei},
  journal={IEEE Transactions on Geoscience and Remote Sensing},
  year={2024},
  publisher={IEEE}
}

@article{2,
  title={Evaluating tool-augmented agents in remote sensing platforms},
  author={Singh, Simranjit and Fore, Michael and Stamoulis, Dimitrios},
  journal={arXiv preprint arXiv:2405.00709},
  year={2024}
}

@article{rs-agent,
  title={RS-Agent: Automating Remote Sensing Tasks through Intelligent Agent},
  author={Xu, Wenjia and Yu, Zijian and Mu, Boyang and Wei, Zhiwei and Zhang, Yuanben and Li, Guangzuo and Peng, Mugen},
  journal={arXiv preprint arXiv:2406.07089},
  year={2024}
}

@article{dmist,
  title={Towards dense moving infrared small target detection: New datasets and baseline},
  author={Chen, Shengjia and Ji, Luping and Zhu, Sicheng and Ye, Mao and Ren, Haohao and Sang, Yongsheng},
  journal={IEEE Transactions on Geoscience and Remote Sensing},
  year={2024},
  publisher={IEEE}
}

@article{ogsod,
  title={Category-oriented localization distillation for sar object detection and a unified benchmark},
  author={Wang, Chao and Ruan, Rui and Zhao, Zhicheng and Li, Chenglong and Tang, Jin},
  journal={IEEE Transactions on Geoscience and Remote Sensing},
  volume={61},
  pages={1--14},
  year={2023},
  publisher={IEEE}
}

@inproceedings{singh2024geollm,
  title={Geollm-engine: A realistic environment for building geospatial copilots},
  author={Singh, Simranjit and Fore, Michael and Stamoulis, Dimitrios},
  booktitle={Proceedings of the IEEE/CVF Conference on Computer Vision and Pattern Recognition},
  pages={585--594},
  year={2024}
}

@article{du2023tree,
  title={Tree-gpt: Modular large language model expert system for forest remote sensing image understanding and interactive analysis},
  author={Du, Siqi and Tang, Shengjun and Wang, Weixi and Li, Xiaoming and Guo, Renzhong},
  journal={arXiv preprint arXiv:2310.04698},
  year={2023}
}

@article{UnivEARTH,
  title={Towards llm agents for earth observation},
  author={Kao, Chia Hsiang and Zhao, Wenting and Revankar, Shreelekha and Speas, Samuel and Bhagat, Snehal and Datta, Rajeev and Phoo, Cheng Perng and Mall, Utkarsh and Vondrick, Carl and Bala, Kavita and others},
  journal={arXiv preprint arXiv:2504.12110},
  year={2025}
}

@inproceedings{yao2025remotesam,
  title={Remotesam: Towards segment anything for earth observation},
  author={Yao, Liang and Liu, Fan and Chen, Delong and Zhang, Chuanyi and Wang, Yijun and Chen, Ziyun and Xu, Wei and Di, Shimin and Zheng, Yuhui},
  booktitle={Proceedings of the 33rd ACM International Conference on Multimedia},
  pages={3027--3036},
  year={2025}
}

@ARTICLE{sar1,
  author={Li, Weijie and Yang, Wei and Hou, Yuenan and Liu, Li and Liu, Yongxiang and Li, Xiang},
  journal={IEEE Transactions on Image Processing}, 
  title={SARATR-X: Toward Building a Foundation Model for SAR Target Recognition}, 
  year={2025},
  volume={34},
  number={},
  pages={869-884},
  doi={10.1109/TIP.2025.3531988}}

@ARTICLE{sar2,
  title = {Predicting gradient is better: Exploring self-supervised learning for SAR ATR with a joint-embedding predictive architecture},
  journal = {ISPRS Journal of Photogrammetry and Remote Sensing},
  volume = {218},
  pages = {326-338},
  year = {2024},
  issn = {0924-2716},
  doi = {https://doi.org/10.1016/j.isprsjprs.2024.09.013},
  url = {https://www.sciencedirect.com/science/article/pii/S0924271624003514},
  author = {Li, Weijie and Yang, Wei and Liu, Tianpeng and Hou, Yuenan and Li, Yuxuan and Liu, Zhen and Liu, Yongxiang and Liu, Li}}

@article{ir,
  title={Towards dense moving infrared small target detection: New datasets and baseline},
  author={Chen, Shengjia and Ji, Luping and Zhu, Sicheng and Ye, Mao and Ren, Haohao and Sang, Yongsheng},
  journal={IEEE Transactions on Geoscience and Remote Sensing},
  year={2024},
  publisher={IEEE}
}

@article{yang2024harnessing,
  title={Harnessing the power of llms in practice: A survey on chatgpt and beyond},
  author={Yang, Jingfeng and Jin, Hongye and Tang, Ruixiang and Han, Xiaotian and Feng, Qizhang and Jiang, Haoming and Zhong, Shaochen and Yin, Bing and Hu, Xia},
  journal={ACM Transactions on Knowledge Discovery from Data},
  volume={18},
  number={6},
  pages={1--32},
  year={2024},
  publisher={ACM New York, NY}
}

@article{zhou2024larger,
  title={Larger and more instructable language models become less reliable},
  author={Zhou, Lexin and Schellaert, Wout and Mart{\'\i}nez-Plumed, Fernando and Moros-Daval, Yael and Ferri, C{\`e}sar and Hern{\'a}ndez-Orallo, Jos{\'e}},
  journal={Nature},
  volume={634},
  number={8032},
  pages={61--68},
  year={2024},
  publisher={Nature Publishing Group UK London}
}

@inproceedings{zhao2024expel,
  title={Expel: Llm agents are experiential learners},
  author={Zhao, Andrew and Huang, Daniel and Xu, Quentin and Lin, Matthieu and Liu, Yong-Jin and Huang, Gao},
  booktitle={Proceedings of the AAAI Conference on Artificial Intelligence},
  volume={38},
  number={17},
  pages={19632--19642},
  year={2024}
}

@inproceedings{li-2025-review,
    title = "A Review of Prominent Paradigms for {LLM}-Based Agents: Tool Use, Planning (Including {RAG}), and Feedback Learning",
    author = "Li, Xinzhe",
    editor = "Rambow, Owen  and
      Wanner, Leo  and
      Apidianaki, Marianna  and
      Al-Khalifa, Hend  and
      Eugenio, Barbara Di  and
      Schockaert, Steven",
    booktitle = "Proceedings of the 31st International Conference on Computational Linguistics",
    month = jan,
    year = "2025",
    address = "Abu Dhabi, UAE",
    publisher = "Association for Computational Linguistics",
    url = "https://aclanthology.org/2025.coling-main.652/",
    pages = "9760--9779",
}

@inproceedings{guo2024remote,
  title={Remote sensing chatgpt: Solving remote sensing tasks with chatgpt and visual models},
  author={Guo, Haonan and Su, Xin and Wu, Chen and Du, Bo and Zhang, Liangpei and Li, Deren},
  booktitle={IGARSS 2024-2024 IEEE International Geoscience and Remote Sensing Symposium},
  pages={11474--11478},
  year={2024},
  organization={IEEE}
}

@article{oubennaceur2019flood,
  title={Flood risk mapping for direct damage to residential buildings in Quebec, Canada},
  author={Oubennaceur, Khalid and Chokmani, Karem and Nastev, Miroslav and Lhissou, Rachid and El Alem, Anas},
  journal={International journal of disaster risk reduction},
  volume={33},
  pages={44--54},
  year={2019},
  publisher={Elsevier}
}

@article{fu2022critical,
  title={Critical role of irrigation efficiency for cropland expansion in western China arid agroecosystems},
  author={Fu, Jianyu and Wang, Weiguang and Zaitchik, Benjamin and Nie, Wanshu and Fei, Esther Xu and Miller, Scot M and Harman, Ciaran J},
  journal={Earth's Future},
  volume={10},
  number={9},
  pages={e2022EF002955},
  year={2022},
  publisher={Wiley Online Library}
}

@article{hansen2009quantifying,
  title={Quantifying changes in the rates of forest clearing in Indonesia from 1990 to 2005 usingremotely sensed data sets},
  author={Hansen, Matthew C and Stehman, Stephen V and Potapov, Peter V and Arunarwati, Belinda and Stolle, Fred and Pittman, Kyle},
  journal={Environmental Research Letters},
  volume={4},
  number={3},
  pages={034001},
  year={2009},
  publisher={IOP Publishing}
}

@inproceedings{wang2021loveda,
  title     = {LoveDA: A Remote Sensing Land-Cover Dataset for Domain Adaptive Semantic Segmentation},
  author    = {Junjue Wang and Zhuo Zheng and Ailong Ma and Xiaoyan Lu and Yanfei Zhong},
  booktitle = {Proceedings of the NeurIPS Track on Datasets and Benchmarks},
  year      = {2021},
  volume    = {1},
}

@article{potsdam,
  title={Semantic segmentation of remote-sensing imagery using heterogeneous big data: International society for photogrammetry and remote sensing potsdam and cityscape datasets},
  author={Song, Ahram and Kim, Yongil},
  journal={ISPRS International Journal of Geo-Information},
  volume={9},
  number={10},
  pages={601},
  year={2020},
  publisher={MDPI}
}

@article{twohig2018health,
  title={The health benefits of the great outdoors: A systematic review and meta-analysis of greenspace exposure and health outcomes},
  author={Twohig-Bennett, Caoimhe and Jones, Andy},
  journal={Environmental research},
  volume={166},
  pages={628--637},
  year={2018},
  publisher={Elsevier}
}

@article{20,
  title={Urban greening to cool towns and cities: A systematic review of the empirical evidence},
  author={Bowler, Diana E and Buyung-Ali, Lisette and Knight, Teri M and Pullin, Andrew S},
  journal={Landscape and urban planning},
  volume={97},
  number={3},
  pages={147--155},
  year={2010},
  publisher={Elsevier}
}

@article{30,
  title={Urban green space cooling effect in cities},
  author={Aram, Farshid and Garc{\'\i}a, Ester Higueras and Solgi, Ebrahim and Mansournia, Soran},
  journal={Heliyon},
  volume={5},
  number={4},
  year={2019},
  publisher={Elsevier}
}

@article{50,
  title={Planning for cooler cities: A framework to prioritise green infrastructure to mitigate high temperatures in urban landscapes},
  author={Norton, Briony A and Coutts, Andrew M and Livesley, Stephen J and Harris, Richard J and Hunter, Annie M and Williams, Nicholas SG},
  journal={Landscape and urban planning},
  volume={134},
  pages={127--138},
  year={2015},
  publisher={Elsevier}
}

@article{r50,
  title={A complex landscape of inequity in access to urban parks: A literature review},
  author={Rigolon, Alessandro},
  journal={Landscape and urban planning},
  volume={153},
  pages={160--169},
  year={2016},
  publisher={Elsevier}
}

@article{51,
  title={Land Cover Classification System (LCCS): classification concepts and user manual},
  author={Di Gregorio, Antonio and Jansen, Louisa JM},
  journal={FAO, Rome},
  year={1998}
}

@article{52,
  title={Object-Based Similarity Assessment Using Land Cover Meta-Language (LCML): Concept, Challenges, and Implementation},
  author={Mosca, Nicola and Di Gregorio, Antonio and Henry, Matieu and Jalal, Rashed and Blonda, Palma},
  journal={IEEE Journal of Selected Topics in Applied Earth Observations and Remote Sensing},
  volume={13},
  pages={3790--3805},
  year={2020},
  publisher={IEEE}
}

@article{s1,
  title={Simulation study of the IAMSAR standard recovery maneuvers for the improvement of serviceability},
  author={Kim, Inchul and Chae, Chongju and Lee, Soyeong},
  journal={Journal of Marine Science and Engineering},
  volume={8},
  number={6},
  pages={445},
  year={2020},
  publisher={MDPI}
}

@article{safety,
  title={Safety distances for storage tanks to prevent fire damage in Wildland-Industrial Interface},
  author={Ricci, Federica and Scarponi, Giordano Emrys and Pastor, Elsa and Planas, Eul{\`a}lia and Cozzani, Valerio},
  journal={Process Safety and Environmental Protection},
  volume={147},
  pages={693--702},
  year={2021},
  publisher={Elsevier}
}

@article{safety2,
  title={Research on the safety and security distance of above-ground liquefied gas storage tanks and dispensers},
  author={Kukfisz, Bo{\.z}ena and Kuczy{\'n}ska, Aneta and Piec, Robert and Szyku{\l}a-Piec, Barbara},
  journal={International journal of environmental research and public health},
  volume={19},
  number={2},
  pages={839},
  year={2022},
  publisher={MDPI}
}

@article{gpt-3.5,
  title={Training language models to follow instructions with human feedback},
  author={Ouyang, Long and Wu, Jeffrey and Jiang, Xu and Almeida, Diogo and Wainwright, Carroll and Mishkin, Pamela and Zhang, Chong and Agarwal, Sandhini and Slama, Katarina and Ray, Alex and others},
  journal={Advances in neural information processing systems},
  volume={35},
  pages={27730--27744},
  year={2022}
}

@article{4o,
  title={Gpt-4o system card},
  author={Hurst, Aaron and Lerer, Adam and Goucher, Adam P and Perelman, Adam and Ramesh, Aditya and Clark, Aidan and Ostrow, AJ and Welihinda, Akila and Hayes, Alan and Radford, Alec and others},
  journal={arXiv preprint arXiv:2410.21276},
  year={2024}
}

@article{team2023gemini,
  title={Gemini: a family of highly capable multimodal models},
  author={Team, Gemini and Anil, Rohan and Borgeaud, Sebastian and Alayrac, Jean-Baptiste and Yu, Jiahui and Soricut, Radu and Schalkwyk, Johan and Dai, Andrew M and Hauth, Anja and Millican, Katie and others},
  journal={arXiv preprint arXiv:2312.11805},
  year={2023}
}

@article{bi2024deepseek,
  title={Deepseek llm: Scaling open-source language models with longtermism},
  author={Bi, Xiao and Chen, Deli and Chen, Guanting and Chen, Shanhuang and Dai, Damai and Deng, Chengqi and Ding, Honghui and Dong, Kai and Du, Qiushi and Fu, Zhe and others},
  journal={arXiv preprint arXiv:2401.02954},
  year={2024}
}

@article{team2024qwen2,
  title={Qwen2 technical report},
  author={Team, Qwen},
  journal={arXiv preprint arXiv:2407.10671},
  volume={2},
  year={2024}
}

@inproceedings{hrscd,
  title={Urban change detection for multispectral earth observation using convolutional neural networks},
  author={Daudt, Rodrigo Caye and Le Saux, Bertr and Boulch, Alexandre and Gousseau, Yann},
  booktitle={IGARSS 2018-2018 IEEE International Geoscience and Remote Sensing Symposium},
  pages={2115--2118},
  year={2018},
  organization={Ieee}
}

@INPROCEEDINGS{deepglobe,  
author={Y. {Li} and L. {Chen}},  
booktitle={2019 IEEE 5th International Conference on Computer and Communications (ICCC)},   
title={Land Cover Classification for High Resolution Remote Sensing Images with Atrous Convolution and BFS},   year={2019},  
volume={},  
number={},  
pages={1808-1813},}

@misc{isprs_vaihingen,
  title        = {ISPRS Vaihingen 2D Semantic Labeling Dataset},
  howpublished = {\url{https://www.isprs.org/resources/datasets/benchmarks/UrbanSemLab/2d-sem-label-vaihingen.aspx}},
  note         = {ISPRS Commission III, WG III/4},
  year         = {2014}
}

@article{qwen7b,
  title={Qwen technical report},
  author={Bai, Jinze and Bai, Shuai and Chu, Yunfei and Cui, Zeyu and Dang, Kai and Deng, Xiaodong and Fan, Yang and Ge, Wenbin and Han, Yu and Huang, Fei and others},
  journal={arXiv preprint arXiv:2309.16609},
  year={2023}
}

@article{InternLM3-8B,
  title={Internlm2 technical report},
  author={Cai, Zheng and Cao, Maosong and Chen, Haojiong and Chen, Kai and Chen, Keyu and Chen, Xin and Chen, Xun and Chen, Zehui and Chen, Zhi and Chu, Pei and others},
  journal={arXiv preprint arXiv:2403.17297},
  year={2024}
}

@article{LLaMA-3-8B,
  title={The llama 3 herd of models},
  author={Dubey, Abhimanyu and Jauhri, Abhinav and Pandey, Abhinav and Kadian, Abhishek and Al-Dahle, Ahmad and Letman, Aiesha and Mathur, Akhil and Schelten, Alan and Yang, Amy and Fan, Angela and others},
  journal={arXiv e-prints},
  pages={arXiv--2407},
  year={2024}
}

@misc{Yi-1.5-6B,
      title={Yi: Open Foundation Models by 01.AI}, 
      author={01. AI and : and Alex Young and Bei Chen and Chao Li and Chengen Huang and Ge Zhang and Guanwei Zhang and Guoyin Wang and Heng Li and Jiangcheng Zhu and Jianqun Chen and Jing Chang and Kaidong Yu and Peng Liu and Qiang Liu and Shawn Yue and Senbin Yang and Shiming Yang and Wen Xie and Wenhao Huang and Xiaohui Hu and Xiaoyi Ren and Xinyao Niu and Pengcheng Nie and Yanpeng Li and Yuchi Xu and Yudong Liu and Yue Wang and Yuxuan Cai and Zhenyu Gu and Zhiyuan Liu and Zonghong Dai},
      year={2025},
      eprint={2403.04652},
      archivePrefix={arXiv},
      primaryClass={cs.CL},
      url={https://arxiv.org/abs/2403.04652}, 
}

@article{abdin2024phi,
  title={Phi-4 technical report},
  author={Abdin, Marah and Aneja, Jyoti and Behl, Harkirat and Bubeck, S{\'e}bastien and Eldan, Ronen and Gunasekar, Suriya and Harrison, Michael and Hewett, Russell J and Javaheripi, Mojan and Kauffmann, Piero and others},
  journal={arXiv preprint arXiv:2412.08905},
  year={2024}
}

@inproceedings{change3d,
  title={Change3D: Revisiting Change Detection and Captioning from A Video Modeling Perspective},
  author={Zhu, Duowang and Huang, Xiaohu and Huang, Haiyan and Zhou, Hao and Shao, Zhenfeng},
  booktitle={Proceedings of the Computer Vision and Pattern Recognition Conference},
  pages={24011--24022},
  year={2025}
}

@misc{earthdial,
      title={EarthDial: Turning Multi-sensory Earth Observations to Interactive Dialogues}, 
      author={Sagar Soni and Akshay Dudhane and Hiyam Debary and Mustansar Fiaz and Muhammad Akhtar Munir and Muhammad Sohail Danish and Paolo Fraccaro and Campbell D Watson and Levente J Klein and Fahad Shahbaz Khan and Salman Khan},
      year={2025},
      eprint={2412.15190},
      archivePrefix={arXiv},
      primaryClass={cs.CV},
      url={https://arxiv.org/abs/2412.15190}, 
}

@misc{geochat,
      title={GeoChat: Grounded Large Vision-Language Model for Remote Sensing}, 
      author={Kartik Kuckreja and Muhammad Sohail Danish and Muzammal Naseer and Abhijit Das and Salman Khan and Fahad Shahbaz Khan},
      year={2023},
      eprint={2311.15826},
      archivePrefix={arXiv},
      primaryClass={cs.CV},
      url={https://arxiv.org/abs/2311.15826}, 
}

@misc{geopix,
      title={GeoPix: Multi-Modal Large Language Model for Pixel-level Image Understanding in Remote Sensing}, 
      author={Ruizhe Ou and Yuan Hu and Fan Zhang and Jiaxin Chen and Yu Liu},
      year={2025},
      eprint={2501.06828},
      archivePrefix={arXiv},
      primaryClass={cs.CV},
      url={https://arxiv.org/abs/2501.06828}, 
}

@misc{earthagent,
      title={Earth-Agent: Unlocking the Full Landscape of Earth Observation with Agents}, 
      author={Peilin Feng and Zhutao Lv and Junyan Ye and Xiaolei Wang and Xinjie Huo and Jinhua Yu and Wanghan Xu and Wenlong Zhang and Lei Bai and Conghui He and Weijia Li},
      year={2025},
      eprint={2509.23141},
      archivePrefix={arXiv},
      primaryClass={cs.CV},
      url={https://arxiv.org/abs/2509.23141}, 
}

@article{opencompass,
  title={OpenCompass: A universal evaluation platform for foundation models, 2023},
  author={Contributors, OpenCompass},
  journal={URL https://github. com/open-compass/opencompass},
  volume={6},
  year={2023}
}

@article{judge1,
  title={Judging llm-as-a-judge with mt-bench and chatbot arena},
  author={Zheng, Lianmin and Chiang, Wei-Lin and Sheng, Ying and Zhuang, Siyuan and Wu, Zhanghao and Zhuang, Yonghao and Lin, Zi and Li, Zhuohan and Li, Dacheng and Xing, Eric and others},
  journal={Advances in neural information processing systems},
  volume={36},
  pages={46595--46623},
  year={2023}
}

@misc{judge2,
      title={G-Eval: NLG Evaluation using GPT-4 with Better Human Alignment}, 
      author={Yang Liu and Dan Iter and Yichong Xu and Shuohang Wang and Ruochen Xu and Chenguang Zhu},
      year={2023},
      eprint={2303.16634},
      archivePrefix={arXiv},
      primaryClass={cs.CL},
      url={https://arxiv.org/abs/2303.16634}, 
}

@article{judge3,
  title={A survey on llm-as-a-judge},
  author={Gu, Jiawei and Jiang, Xuhui and Shi, Zhichao and Tan, Hexiang and Zhai, Xuehao and Xu, Chengjin and Li, Wei and Shen, Yinghan and Ma, Shengjie and Liu, Honghao and others},
  journal={The Innovation},
  year={2024},
  publisher={Elsevier}
}

@inproceedings{earthdial1,
  title={Earthdial: Turning multi-sensory earth observations to interactive dialogues},
  author={Soni, Sagar and Dudhane, Akshay and Debary, Hiyam and Fiaz, Mustansar and Munir, Muhammad Akhtar and Danish, Muhammad Sohail and Fraccaro, Paolo and Watson, Campbell D and Klein, Levente J and Khan, Fahad Shahbaz and others},
  booktitle={Proceedings of the Computer Vision and Pattern Recognition Conference},
  pages={14303--14313},
  year={2025}
}

@inproceedings{geochat1,
  title={Geochat: Grounded large vision-language model for remote sensing},
  author={Kuckreja, Kartik and Danish, Muhammad Sohail and Naseer, Muzammal and Das, Abhijit and Khan, Salman and Khan, Fahad Shahbaz},
  booktitle={Proceedings of the IEEE/CVF Conference on Computer Vision and Pattern Recognition},
  pages={27831--27840},
  year={2024}
}

@article{geopix1,
  title={GeoPix: A multimodal large language model for pixel-level image understanding in remote sensing},
  author={Ou, Ruizhe and Hu, Yuan and Zhang, Fan and Chen, Jiaxin and Liu, Yu},
  journal={IEEE Geoscience and Remote Sensing Magazine},
  year={2025},
  publisher={IEEE}
}

\newpage
\appendix

\end{document}